\documentclass[conference]{IEEEtran}

\usepackage{times}
\usepackage{epsfig}
\usepackage{graphicx}
\usepackage{amsmath}
\usepackage{amssymb}
\usepackage{lipsum}  
\usepackage{multirow}
\usepackage{url}
\usepackage{amssymb}
\usepackage{graphicx}

\begin{document}
%
\title{A Parameterized Approach to Personalized Variable Length Summarization of Soccer Matches}
\author{\IEEEauthorblockN{Mohak Sukhwani}
\IEEEauthorblockA{IBM Research\\ Bangalore, India\\mosukhwa@in.ibm.com}
\and
\IEEEauthorblockN{Ravi Kothari}
\IEEEauthorblockA{IBM Research\\ Bangalore, India\\rkothari@in.ibm.com}
}

\maketitle

\begin{abstract}
In this paper, we present a parameterized approach to produce personalized variable length summaries of soccer matches. Our approach is based on temporally segmenting the soccer video into ``plays'', associating a user-specifiable ``utility'' for each type of play and using ``bin-packing'' to select a subset of the plays that add up to the desired length while maximizing the overall utility (volume in bin-packing terms). Our approach systematically allows a user to override the default weights assigned to each type of play with individual preferences and thus see a highly personalized variable length summarization of soccer matches. We demonstrate our approach based on the output of an end-to-end pipeline that we are building to produce such summaries. Though aspects of the overall end-to-end pipeline are human assisted at present, the results clearly show that the proposed approach is capable of producing semantically meaningful and compelling summaries. Besides the obvious use of producing summaries of superior league matches for news broadcasts, we anticipate our work to promote greater awareness of the local matches and junior leagues by producing consumable summaries of them. 
\end{abstract}


%
\IEEEpeerreviewmaketitle

\section{Introduction}

\begin{figure}[t]
\begin{center}
   \includegraphics[width=0.95\linewidth]{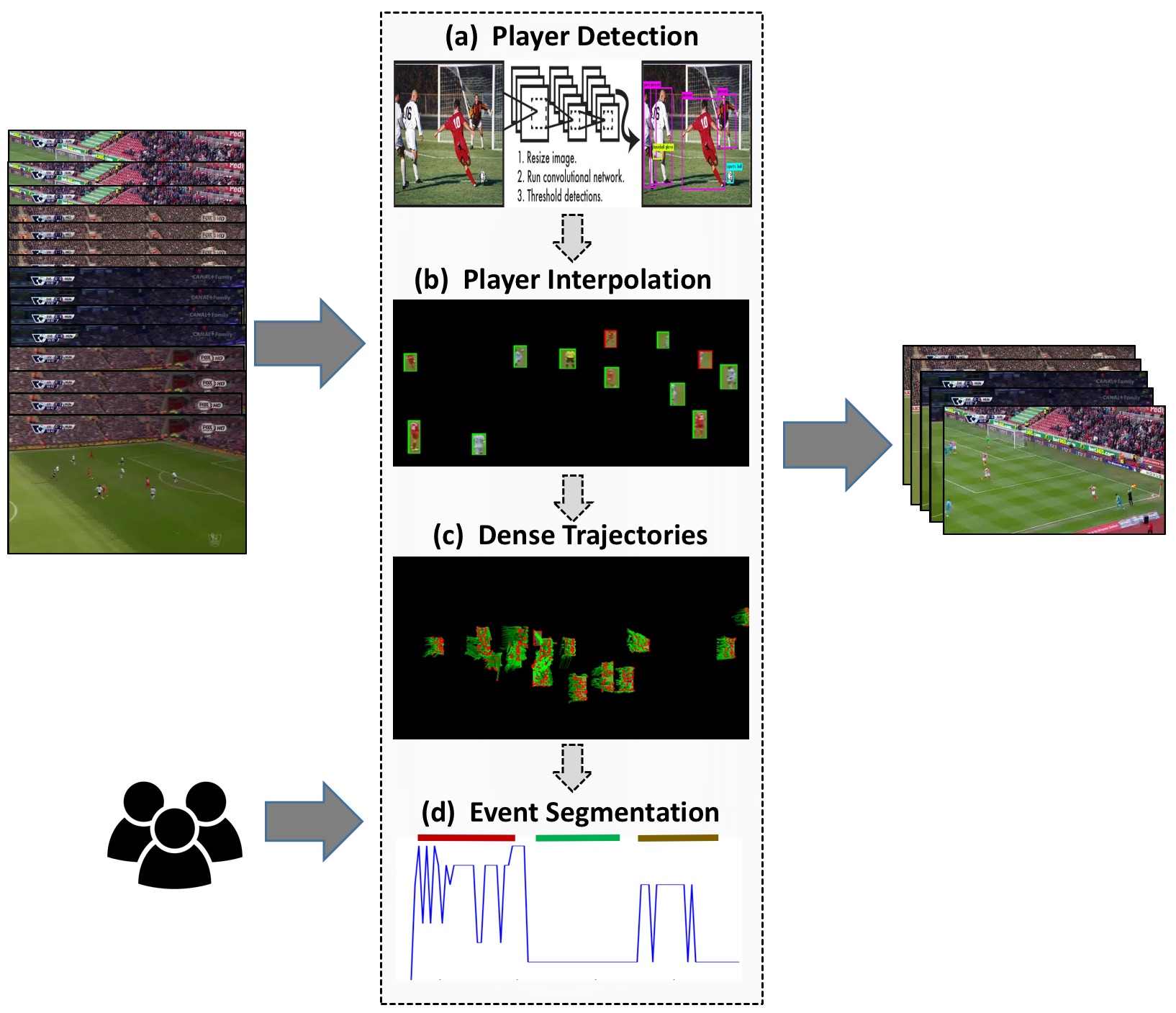}
\end{center}
\caption{The proposed system generates variable length video summaries for the game of soccer. The summary is generated by temporally segmenting the soccer video into ``plays'', associating a user-specifiable ``utility'' for each type of play, and using ``bin-packing'' to select a subset of the plays to add up to the desired length while maximizing the overall utility. The summarization block is composed of a player localization module, an event segmentation module and an event selection module.}
\label{fig:abstract}
\end{figure}

\begin{figure*}[t]
\begin{center}
\includegraphics[width=0.95\linewidth]{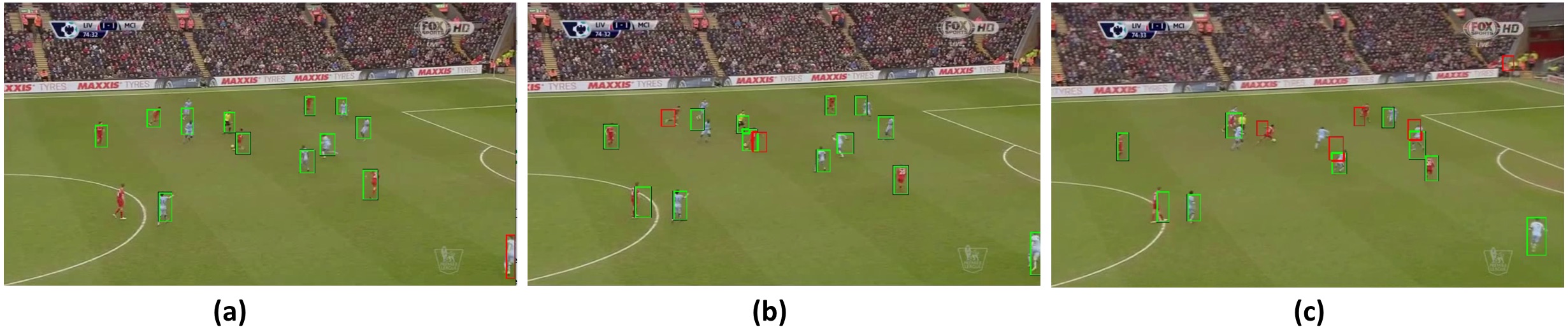}
\end{center}
   \caption{Player interpolation to compensate missing detections. Red boxes are obtained through frame interpolation. Various cases of bounding box (frame) interpolation -- (a) Three neighbouring frames (b) Five neighbouring frame (c) Seven neighbouring frame. One can notice a surge in false detections with increasing neighbourhood size. Interpolation with $3$ neighbourhood frames is best suited for the proposed approach.}
   
\label{fig:playerRecog}
\end{figure*}

Summarization is the representation of a long video by a compact sequence. Broadly, there are two ways of summarizing a video sequence. \textit{Key frame} based summarization stitches together individual frames, that together, are supposed to be representative of the overall video. \textit{Video skimming} on the other hand, identifies sub-sequences (``clips'') to create the overall summary. Each clip may contain a few or several hundred contiguous frames and thus encapsulates the context as well as the audio required for a more thorough understanding of the video. Summarization thus allows a user to rapidly assimilate the contents of the overall video with a relatively small investment of time. Summarization can also enable storage, complex indexing and retrieval while optimizing the overall resource consumption. 

Summarization is subjective -- what is important to one user may be less important to another. For example, a \textit{goal keeper} may prefer a summarization that emphasizes the activity near the goal while a \textit{center forward} may prefer a summarization that is biased towards attacking plays. A fan may prefer a summarization that shows more footage of a specific player and so on. We thus bring in the flavour of human centered design with focus on individualistic needs.

Our work is motivated by two separate considerations. On one hand, we aim to incorporate a user's perspective to drive the summarization. On the other hand, we want to create variable length summaries in an objective manner. Our key contributions are, (i) the use of change point of visual features to decide segmentation points and hence plays or clips, (ii) representing each clip using optical flow, (iii) an optional granular and systematic way of allowing a user to specify a ``utility'' for each type of play, and (iv) an optimization driven approach to produce arbitrary length summarizations. We have started constructing an end-to-end pipeline that creates variable length summaries based on user specified preferences. Our pipeline, Figure~\ref{fig:abstract}, has steps which at present require human assistance; however, the summarization aspects that we describe in this paper do remarkably well with minimal assistance. We believe our work is more generic though we focus exclusively on \textit{soccer} in this paper. 

We have laid out the rest of the paper as follows. In Section~\ref{Section:Review}, we provide a brief summary of the motivation and related work. In Section~\ref{Section:Overall}, we provide an outline of our approach and establish some definitions that allows for segmentation of video sequences into semantically coherent clips. In penultimate Section~\ref{Section:Results}, we describe the experimental setup and discuss the results obtained. We end with conclusions and future work in Section~\ref{Section:Conclusion}. 

\section{Related Work} \label{Section:Review}

One of the simplest approaches to key frame based segmentation is based on clustering all the frames and then choosing the frame that is best representative of the cluster~\cite{Sony11,Hadi06}. Enhancement can include clustering in a perceptual space (i.e. the distance measure used is perceptually motivated~\cite{Li03}). Other approaches to key frame based summarization are based on detecting changes in features~\cite{Zhang97,Ciocca06} and motion activity ~\cite{Wolf96,Liu03,Congcong09}. Sudden change in the audio levels can also be a basis for key frame selection. However, as mentioned, key frame summarization methods select individual frames and thereby loose the context in which that frame occurred. For example, a frame containing a circle could have a triangle gradually morphing into a circle; however that evolution is lost if only the frame containing the circle is retained in the summarization. In addition, the loss of the audio track only adds to the coarseness of the summarization. 

One of the early techniques for video skimming relies on language understanding along with visual features \cite{Smith97}. However, the reliance on language understanding constrains the applicability to where meaningful and discernible audio is present. Subsequently, EM \cite{Orriols01}, and SVD \cite{Gong00} were also applied. A user attention based approach for video summarization \cite{Ma02} models user's attention and then selects the frame or a sequence of frames (clips) that are most likely to attract attention (as predicted by the model). More recently, there is a growing interest in detecting primitives (such as shots \cite{Cotsaces06}, events \cite{Li13}) and using that as the basis for summarization. However, the drawback of being able to systematically parameterize summarization while being able to produce summaries of variable lengths remains. 

Sports video summarization in general has seen considerable traction in recent times -- player and ball motion tracking~\cite{maksai2016players}, highlight generation using audio visual cues ~\cite{li2006soccer,ye2010audio} etc. are being actively pursued in constrained settings. A recent and an interesting approach to generate basketball highlights~\cite{bettadapura2016leveraging} leverages multimodal context cues for video summarization. While the system asserts good results in indoor basketball settings, replicating same to the outdoor settings would be a challenge in the absence of high quality broadcast data complemented with player and play-by-play stats. Local soccer matches or junior leagues are devoid of such elaborate setups and we only have video data in such settings. We thus propose a generic parameterized approach based on `visual motion' cues to generate highlights of the soccer matches.

\section{Formulation and Analysis} \label{Section:Overall}

\begin{figure}[b]
\begin{center}
  \includegraphics[width=0.95\linewidth]{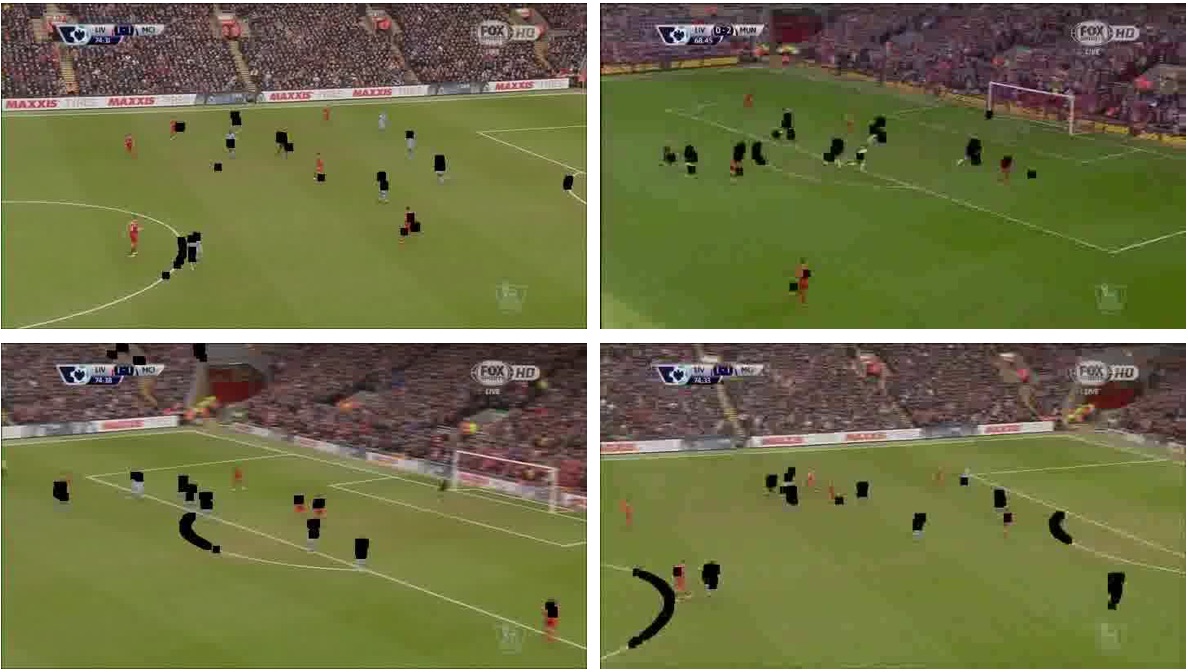}
\end{center}
\caption{Instances of detections with background subtraction using GMM.}
\label{fig:playerGMM}
\end{figure}

Our proposed approach is based on defining segmentation points as change points of the visual features. The segmentation points, for example, are points that include (but are not limited to) regions where, (i) the optical flow shows a significant change e.g. multiplicity of players moving in one direction and then suddenly moving in the opposite direction, (ii) the optical flow becomes laminar, (iii) velocity of players decays to $0$ e.g. the ball going out of the sidelines results in the velocity of a multiplicity of players approaching $0$ (albeit, for a short duration). We provide a detailed description of each of the building blocks of our pipeline in the upcoming sections. 

\subsection{Player Detection}

Current person detection systems use classifiers over Histogram of Oriented Gradients (HOG)~\cite{hog} or Deformable Parts Models (DPM)~\cite{dpm} for detecting humans in an image sequence. To localize the person(s), both HOG and DPM based systems generally operate in a sliding window fashion which makes them very inefficient. Moreover, their accuracy is also marred by size of the players in the soccer field (a player in soccer field occupies a few pixels). We use the following two methods to instantly locate players in the moving frames.  

\subsubsection{Background Subtraction using GMM}

We use a Gaussian Mixture Model (GMM) to model a soccer field and use the following steps to detect the players on the soccer field.

\begin{itemize}

\item~\textit{Isolate Playing Field}: We take patches of the field and model it using a fixed number of Gaussians. For simplicity, the last $5$ rows of the frames are selected as field pixels to train the GMM -- owing to the choice of camera angle in our dataset, this assumption is true everytime -- the lower part of the frames would correspond to field pixels). We get a two-fold advantage with such a design choice: (i) there is no need to manually curate field patches, and (ii) the mixture can be adapted to incorporate the subtleties of a specific field design, light variations etc. We deploy per pixel classification using learned GMM to segment the playing field -- the foreground pixels represent the soccer field.~\footnote{The Gaussian learning is an online process and the parameters are regularly updated. This makes it robust to changes that happen over the time in video, viz. change in light illumination on the field.}

\item~\textit{Detect players}: Potential player regions are extracted by subtracting present frame from a moving average of the past $10$ frames. The moving average computation allows us to capture subtle variations that occur in background over time. The regions which lie outside the ``field markers'' are discarded. Field markers are detected using Hough shape analysis. We perform morphological operations over detected blobs to isolate player patches, Figure~\ref{fig:playerGMM}.

\end{itemize}

\begin{figure}
\begin{center}
\includegraphics[width=\linewidth]{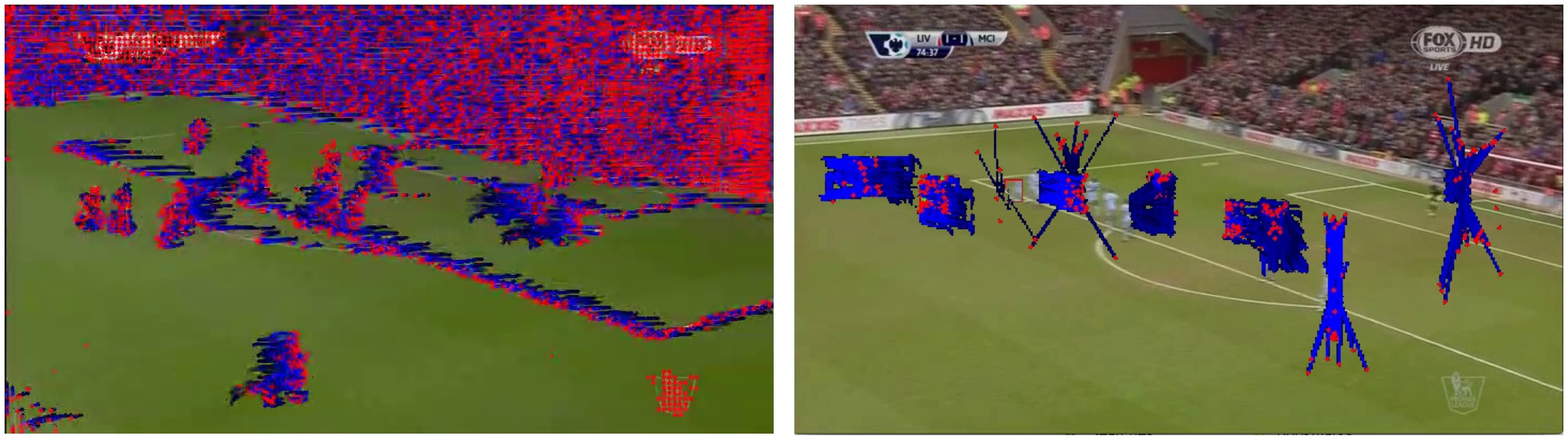}
\end{center}
  \caption{Visualization of the motion trajectories.~\textit{(Left to Right)} (a) On original frames -- camera motion and other unaccounted interest points dominate, resulting in surge of false trjectories (b) Without player bounding box interpolation -- leads to burst in trajectories at places where players were missed by the `player detector'. We use both player detection and interpolation for smoother trajectory computation and better event segmentation.}
\label{fig:traject}
\end{figure}

\begin{figure}
\begin{center}
\includegraphics[width=\linewidth]{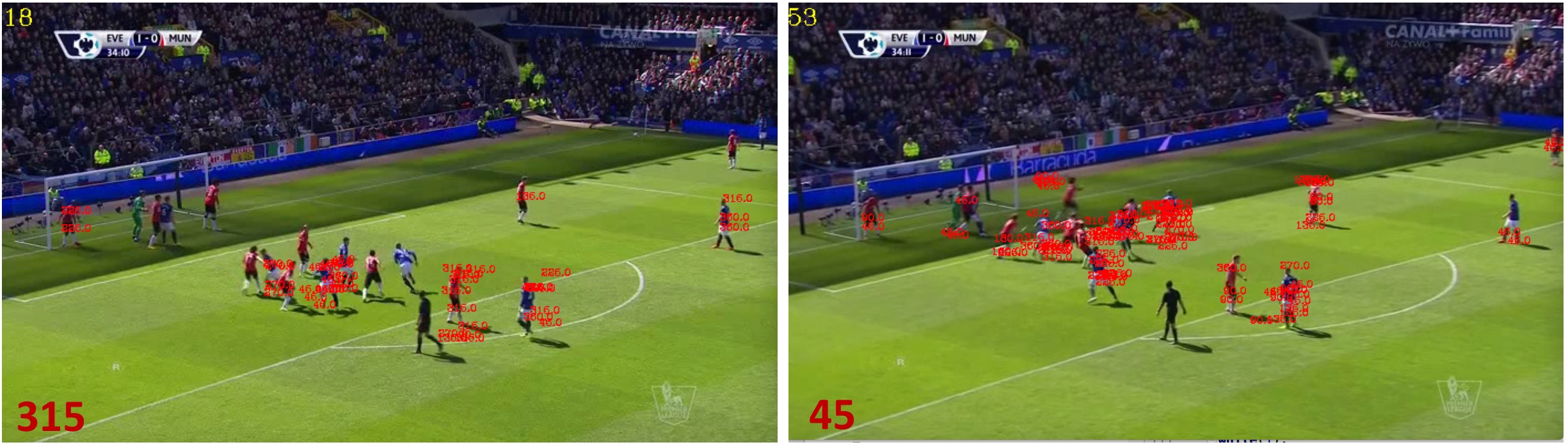}
\end{center}
   \caption{Motion direction (angles) computed using optical flow: Individual player motion directions (clustered numbers near players). Numbers at the top left corner represent frame number. Numbers at the bottom left corner show dominant direction (angle measured counter-clockwise from the horizontal) of overall player motion and represent the motion direction of the frame.}
\label{fig:angle}
\end{figure}

\begin{figure}[b]
\begin{center}
\includegraphics[width=0.9\linewidth]{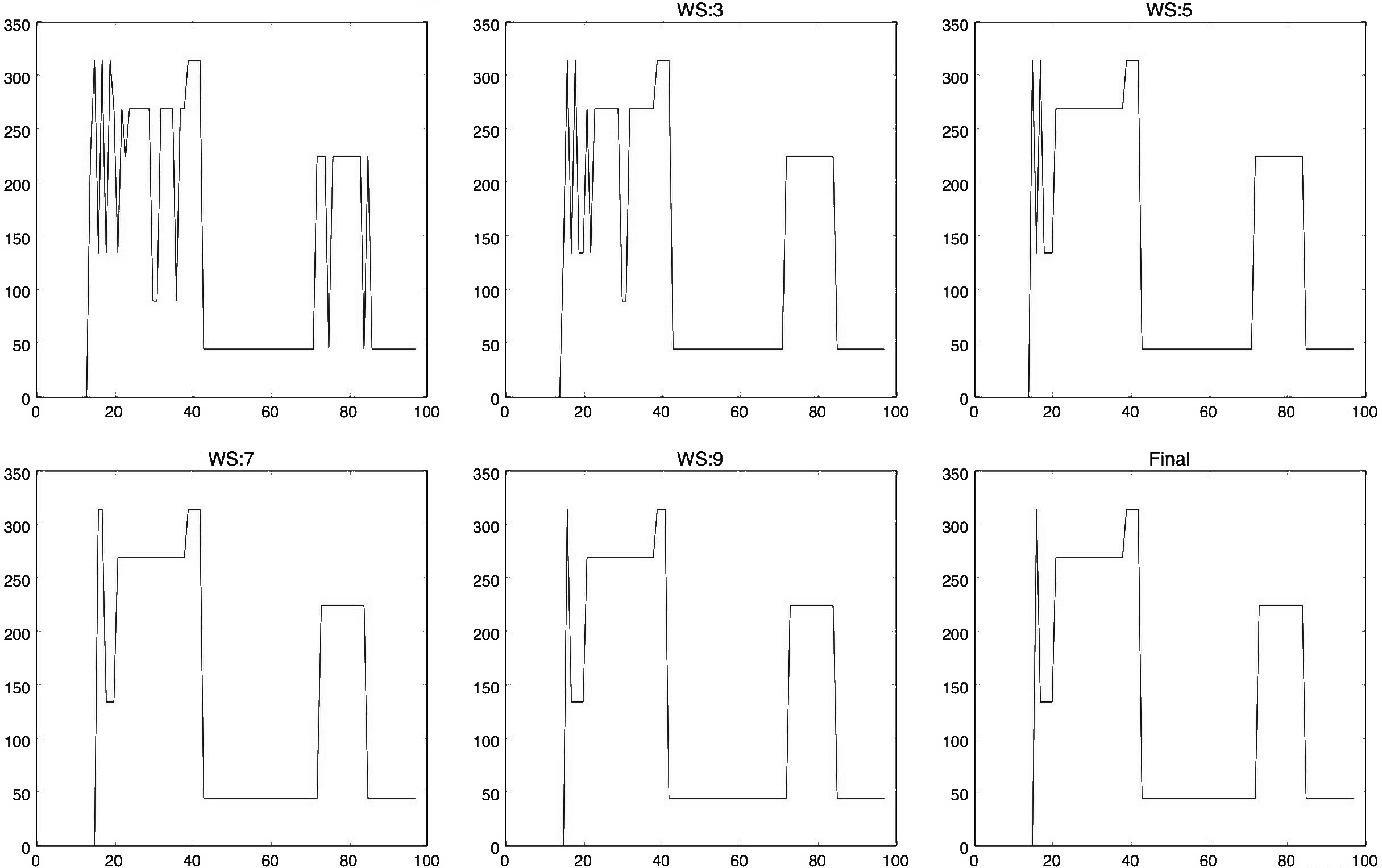}
\end{center}
   \caption{The process of smoothing the ``clips''. The raw signal is convolved with four mode filters with a neighbourhood size of $3,5,7, 9$ (WS: window size) to suppress noisy data points. The final soccer clip representation is computed using the output of convolved representations. ``x axis'' represents the frame number and ``y axis'' represents the angle.}
\label{fig:smooth}
\end{figure}

\subsubsection{Deep Learning for Player Detection}

Recent deep learning based approaches~\cite{fasterRCNN} use region proposal methods to generate potential ``player'' bounding boxes over which classifiers are run for detection. The bounding boxes are then refined and are re-scored to return exact player locations on field. Such a pipeline is hard to optimise and is devoid of real time performance. Player detection is framed as a regression problem using the current state of the art YOLO system ~\cite{yolo}. It uses a single ConvNet to simultaneously predict potential ``player'' bounding boxes and their class probabilities, Figure~\ref{fig:playerRecog}(a). Being extremely fast and accurate the system best suits our need of real time player detection on the field.

False player detections (if any) are pruned out using heuristics - (i) height of the player should be more than width, and (ii) width should be more than $5$ pixels. Failed detections are compensated by propagating player bounding boxes across the neighboring frames, Figure~\ref{fig:playerRecog}. An overlap of less than $25\%$ between bounding boxes in consecutive frames acts as a trigger for player interpolation, Figure~\ref{fig:playerRecog}(b). Each player bounding box is represented as:  

\begin{figure*}[t]
\begin{center}
\includegraphics[width=0.95\linewidth]{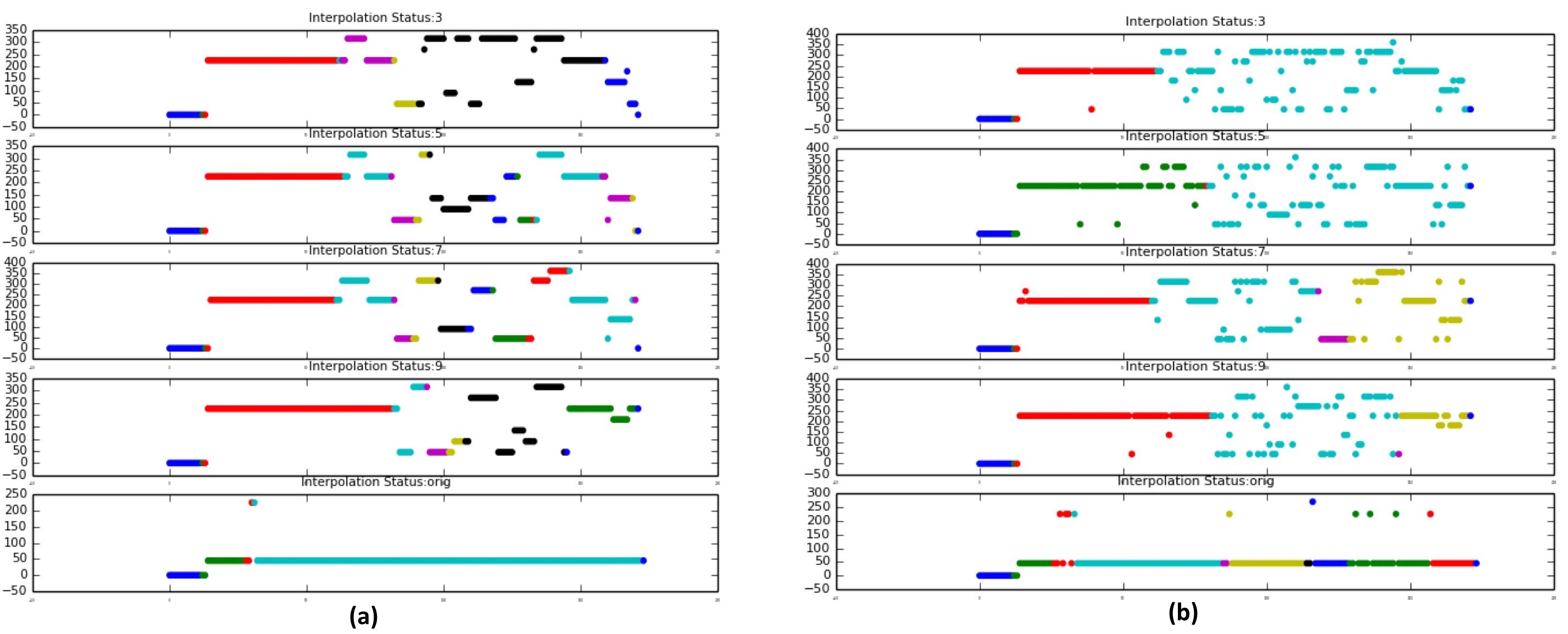}
\end{center}
   \caption{Change point detection and soccer event segmentation on the (a) smoothed representation (b) raw representation. Event segments in the soccer clips are identified using colors. ``Interpolation status'' on the plots in (a) and (b) represent the count of neighbouring frames used to interpolate missing player detections. ``orig'' represent the computation of motion features from full frame without explicit player detection. ``x axis'' represents the frame number and ``y axis'' represents the angle.}
\label{fig:segments}
\end{figure*}

\begin{equation}
     b_i=\begin{bmatrix}
         x_i &  y_i &  w_i & h_i
        \end{bmatrix}
\end{equation}
where $(x_i,y_i)$ are frame co-ordinates and $(w_i,h_i)$ are its corresponding width and height. Given a set of consecutive annotations $b_i$ and $b_{i+1}$, we use the mode of $w_j$'s and $h_j$'s and interpolate the values of $(x_j,y_j)$ using, 
    
\begin{equation}
   \begin{bmatrix} x_j \\ y_j  \end{bmatrix}=  \Bigl( \dfrac{t_{i+1} - t_j}{t_{i+1} - t_i} \Bigr) \begin{bmatrix} x_i \\ y_i \end{bmatrix} + \Bigl( \dfrac{t_{j} - t_i}{t_{i+1} - t_i} \Bigr) \begin{bmatrix} x_{i+1} \\ y_{i+1} \end{bmatrix}
\end{equation}

\subsection{Dense Trajectories For Soccer Events}
We model player motion and activity patterns to isolate soccer events in the given soccer video. Dense Feature trajectory descriptors are used to describe soccer players actions over space-time volume~\cite{impTraject}. The trajectories are densely extracted from multiple spatial scales, Figure~\ref{fig:traject}. They are sampled from a player's region grid space and tracked on each scale separately. The camera motion correction is explicitly handled as a default property of the feature computation. Feature points belonging to non-players regions are pruned out and suppressed to remove any for inconsistencies in soccer event representations. Non-player points tend to add noise and are not real representations of player actions, Figure~\ref{fig:traject}. We compute HOF along the dense trajectories to capture player motions on the field. Each flow vector is binned according to the angle it subtends with the horizontal axis and weighted according to the magnitude. HOF orientations are quantized into $8$ bins using full orientation; an additional bin accounts for count details of flow vectors missing magnitude threshold. The computed descriptors are $L_2$ normalized.

\subsection{Soccer Event Segmentation}

Each frame of the soccer clip is represented by the dominant player motion angle in the frame (Figure~\ref{fig:angle}). A soccer clip thus has a time series representation, with each sample being a dominant player motion angle in a frame, Figure~\ref{fig:angle}. This soccer clip representation is noisy and assimilates various transitions within. We use mode filters to clean up the isolated noise points (Figure~\ref{fig:smooth}). Soccer event segmentation from the filtered soccer clip representation is analogous to change point detection in a time series data -- multiple change points are detected using Bayesian inference~\cite{eventSeg}. Each soccer clip is composed of multiple event segments, Figure~\ref{fig:segments}. We bin these identified events to create final soccer video summary.  

\begin{figure}[hb]
\begin{center}
   \includegraphics[width=0.95\linewidth]{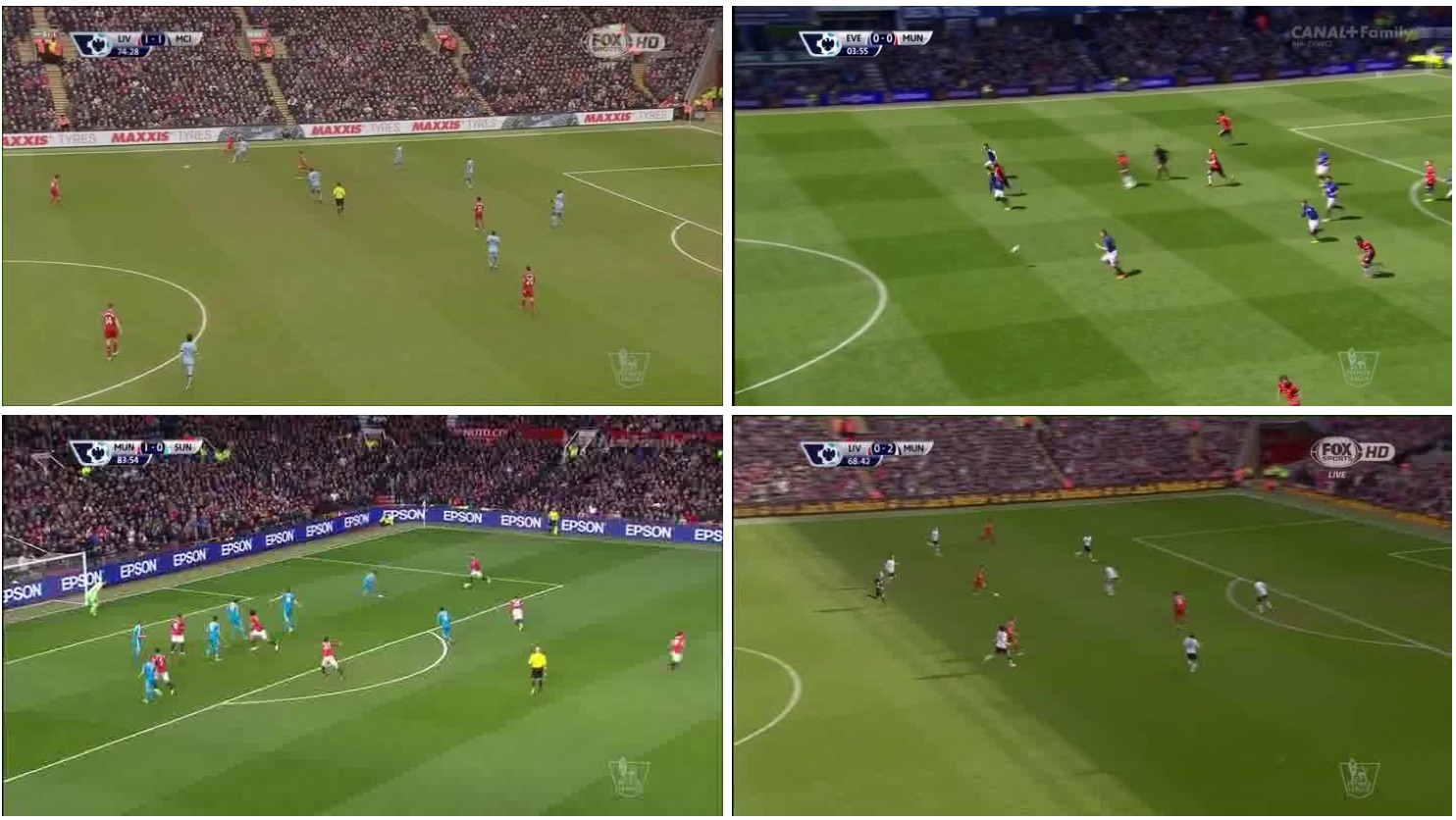}
\end{center}
\caption{Variations in dataset: Instances of illuminations and ground variations in videos from our soccer video dataset.}
\label{fig:dataset}
\end{figure}

\begin{figure*}
\begin{center}
   \includegraphics[width=0.95\linewidth]{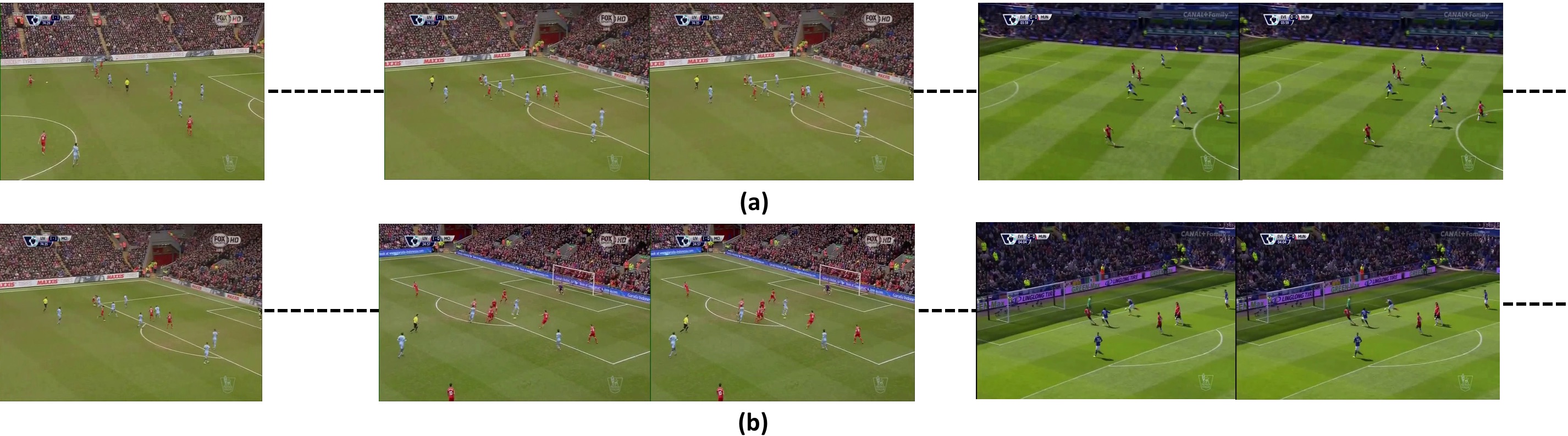}
\end{center}
   \caption{Illustration of variable length summaries of soccer matches generated using the suggested approach. The summaries generated are parameterised over length (time) : (a) 5 sec (b) 10 sec. Both videos are generated at $30$ fps and encapsulate within the different content (observe difference in frame content sampled at same time instance from both the videos).}
\label{fig:opVid}
\end{figure*}

\subsection{Soccer Video Summarisation}

We assign a (pre-defined) utility (weight) to each computed event segment. Once the weights are defined, producing the summarization is a matter of choosing the clips so as to maximize the total utility (weight) of the selected clips while ensuring that the overall length of the summarization does not exceed the specified duration. Our approach is based on bin-packing in which the length of the summarization, $L$, is the size of the bin, the $n$ clips $C_1, C_2, \ldots, C_n$ are items which need to be packed into the bin (knapsack problem). More formally, given a length $L$ (bin size) and $n$ clips (items) of length $C_1, C_2, \ldots, C_n$, maximize, 
\begin{eqnarray}
J & = & \sum_{i=1}^{n} I_i C_i, \quad I_i \in \{0,1\} \\ \nonumber
\textrm{s.t.} & &  \sum_{i=1}^{n} I_i C_i \le L  
\end{eqnarray}

\section{Experiments and Results}\label{Section:Results}

\begin{table}[b]
\begin{center}
\begin{tabular}{|c||c|c|}
\hline
 \small{\bf Time} & \small{\bf Event Preference} & \small{\bf Selected Event List} \\
\hline
\scriptsize{\bf \multirow{4}{3em}{5}} & \scriptsize {[\textit{45}, \textit{0}, 135, 90, 225, 180, 315, 270, 360]} & \scriptsize{[1 6 10 17 20 24 ...]} \\
& \scriptsize {[\textit{135}, \textit{90}, \textit{0}, 180, 225, 45, 315, 360, 270]} & \scriptsize{[16 17 19 26 30 33 35 ...]} \\ 
& \scriptsize {[\textit{270}, \textit{225}, 0, 315, 90, 45, 360, 180, 135]} & \scriptsize{[10 17 28 30 33 35 56 ...]} \\ 
& \scriptsize {[\textit{360}, \textit{270}, \textit{225}, \textit{90}, 45, 0, 180, 135, 315]} & \scriptsize{[30 33 35 50 56 61 67 ...]} \\ 
\hline
\scriptsize{\bf \multirow{4}{3em}{10} } & \scriptsize {[\textit{0}, \textit{45}, \textit{90}, 135, 180, 225, 270, 315, 360]} & \scriptsize{[0 1 6 10 11 17 18 20 .... ]} \\
&\scriptsize {[\textit{90}, \textit{135}, 180, 0, 45, 225, 270, 315, 360]} & \scriptsize{[0 2 10 11 12 16 17 ...]}\\ 
& \scriptsize {[\textit{225}, 270, 315, 0, 45, 90, 135, 180, 360]} & \scriptsize{[5 17 21 25 28 30 33 ... ]}\\ 
& \scriptsize {[\textit{360}, \textit{270}, \textit{225}, 90, 0, 45, 135, 180, 315]} & \scriptsize{[5 10 17 25 30 33 35 ... ]}\\ 
\hline
\end{tabular}
\end{center}
\caption{Quantitative details of personalised soccer summarization parameterized over: (a) Time (sec) and (b) Event Choice. Depending on the parameters we get varied output for each case (output event list varies in each case). The size of the output video is constrained by `Time' parameter and the output video content depends on the selected event list. The~\textit{italicized} numerals represent user explicit selections and others are implicitly determined by the system.}
\label{tab:quanRes}
\end{table}

\begin{figure*}[t]
\begin{center}
   \includegraphics[width=0.95\linewidth]{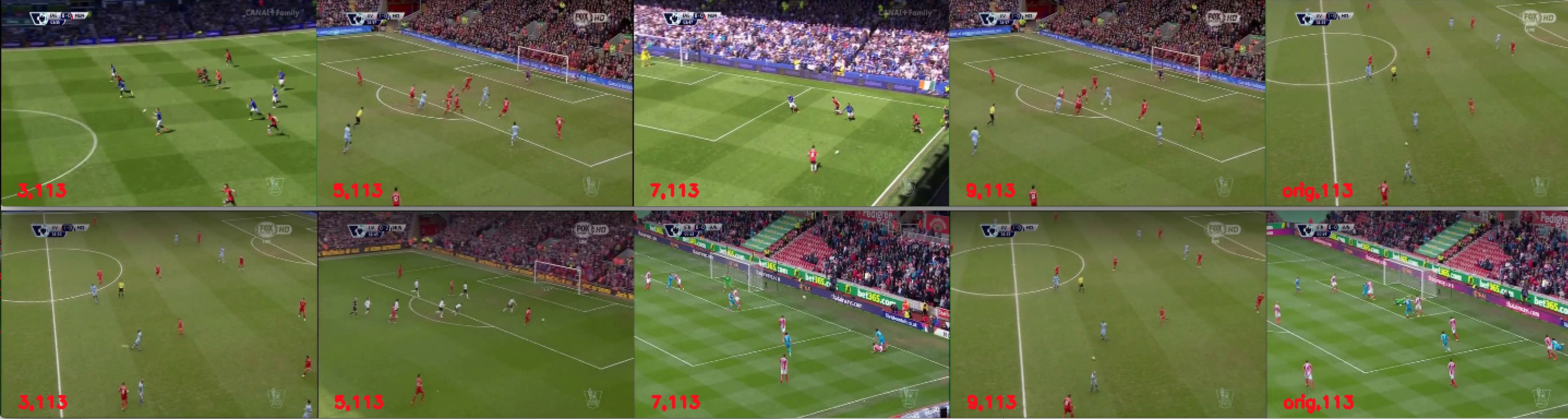}
\end{center}
\caption{Instances of video summaries illustrating variations in content generated. Numbers mentioned at bottom left of the frame represent $<$interpolation-status, frame-number$>$ -- (upper) 10 seconds summary, and (lower) 25 seconds summary.}
\label{fig:compHigh}
\end{figure*}

We use soccer broadcast video recordings with a frame speed of $25$ fps. The soccer video dataset, has an average clip length $249$ frames, is hand curated. Each frame is of resolution $720 \times 400$ and encapsulates the side/longitudinal view of the game play. The video dataset, Figure~\ref{fig:dataset}, is comprised of videos encompassing various variations within.

\textbf{Player Detection and Interpolation}: We test two ways of player detection on the soccer field. Qualitatively fine tuned YOLO performs better than the GMM based method. Background subtraction using GMM does not generalize well to learn ground pixels and returns many false positives (including non-player regions) (Figure~\ref{fig:playerGMM}). We thus use a fine-tuned YOLO ``player detector'' and use a $3$ neighbourhood interpolation to fill in missing player bounding boxes. Interpolation with $3$ neighbourhood frames perform best with an average $12$\% increase in the number of (correct) bounding boxes. Interpolation with $5$, $7$ and $9$ neighbourhood leads to surge in false detections (Figure~\ref{fig:playerRecog}) with average increase of $22$\%, $32$\%, $41$\% respectively in number of bounding boxes. With an average top speed of $28.2$kph~\cite{FIFA}, $3$ frames interpolation at $30$fps translates to tracking player over distance of $\sim1m$, anything beyond this leads to increase in false predictions.     

~\textbf{Soccer Event Segmentation}: Dominant angle computation for soccer video representation using dense trajectories from full frame is influenced by non-player regions (ground pixels, audience pixels, non playing regions etc). This leads to false computations of dominant angle and highly un-even (sudden jerks due to outliers) event segmentation (Figure~\ref{fig:segments}(b) last panel). To avoid the influence of the outliers we smoothen the soccer representation signal and then compute event segments. We generate much smoother and ``fine-grained'' event segments with dense trajectories computed only with player bounding boxes. Soccer video representation computed only with player bounding boxes is a true representative of on-field events on the soccer field and is devoid of any influence of non-playing regions.  

~\textbf{Soccer Video Summarization}: We ensure the ordering of the clips is true to their original order of occurrence making the summaries effective (Figure~\ref{fig:opVid}). For comparison, we illustrate the instances of video summaries using all the methods in Figure~\ref{fig:compHigh}. Visually, illustrated frames might look similar and point to same event, but on careful observation we can look into intrinsic and subtle frame skips ``player bounding box'' methods provide without any qualitative loss. These methods cleverly discard `non-essential' portions of events and thus allow packing much more interesting content in the final output video. `orig' event segments on other hand fail to do this quick transitions due to inherent nature of the segments, Figure~\ref{fig:segments}(a).
 
~\textbf{Summarization as Bin Packing}: The output video (in the present form) is parameterized over time and the user priority list. The user states the preference for the events and the system generates the output summary of the choice, Table~\ref{tab:quanRes}. We assign weights to the user choices in descending order, with maximum value to user\'s first preference. The events which are not explicitly stated by user are randomly selected by system and lower values are assigned them. The constraint remains similar as above, the total length of packed events cannot exceed the knapsack's capacity (time). The goal here is to maximize the total weight (utility) of the selected events so as to produce best possible summary of user choice.

\section{Conclusion and Future Work}\label{Section:Conclusion}

For the future version of the work we would like to conduct extensive studies with more parameters and variations. Each segment of the video sequence could be classified into a set of pre-defined classes \textit{viz.} \textit{Ball Out of Play}, \textit{Throw}, \textit{Attack}, \textit{Goal Kick}, \textit{Corner}, \textit{Free Kick}, \textit{Penalty} etc. could then be assigned pre-computed weights. For example, the weight assigned to a \textit{Penalty Kick} could be $10$ making it almost certain for it to be included in the final summarization. On the other hand, a \textit{Goal Kick} could be assigned a weight of $2$ making it much less likely for it to be included in the final summarization. This essentially would be a better and robust version of the above solution and would give more control to the users to reflect their preferences. We would also like to use other implicit contexts such as audio cues, player identification, player tracks to better parameterize the soccer video summarization process. 

In this paper, we presented an approach to generate highly personalized variable length summarization of the soccer matches. Our approach is equally applicable to generating a summarization of a single match or for complex summarizations from multiple matches (e.g. all goals scored in the world cup). By assigning a utility to specific plays and using a bin packing algorithm to maximize the overall utility, we allow the summarization to be systematic, objective and extensible to arbitrary lengths. Our ongoing work comprises of fully automating the processing pipeline, including several other cues for identifying plays (e.g. audio levels) and transforming the audio to be synchronized to the summarization. 

{
\bibliographystyle{IEEEtran}
\bibliography{soccer}
}

\end{document}